\documentclass{article}

\usepackage{PRIMEarxiv}

\usepackage[utf8]{inputenc} % allow utf-8 input
\usepackage[T1]{fontenc}    % use 8-bit T1 fonts
\usepackage{hyperref}       % hyperlinks
\usepackage{url}            % simple URL typesetting
\usepackage{booktabs}       % professional-quality tables
\usepackage{amsfonts}       % blackboard math symbols
\usepackage{nicefrac}       % compact symbols for 1/2, etc.
\usepackage{microtype}      % microtypography
\usepackage{fancyhdr}       % header
\usepackage{graphicx}       % graphics
\usepackage{multirow}
\usepackage{pifont}
\usepackage{xcolor} 
\graphicspath{{media/}}     % organize your images and other figures under media/ folder

\title{Pushing Trade-Off Boundaries: Compact yet Effective Remote Sensing Change Detection}

\author{
  Luosheng Xu \\
  Space Information Research Institute \\
  Hangzhou Dianzi University \\
  \and
  Dalin Zhang \\
  Space Information Research Institute \\
  Hangzhou Dianzi University \\
  \and
  Zhaohui Song \\
  Space Information Research Institute \\
  Hangzhou Dianzi University \\
}

\begin{document}
\maketitle

\begin{abstract}
Remote sensing change detection is essential for monitoring urban expansion, disaster assessment, and resource management, offering timely, accurate, and large-scale insights into dynamic landscape transformations. While deep learning has revolutionized change detection, the increasing complexity and computational demands of modern models have not necessarily translated into significant accuracy gains. Instead of following this trend, this study explores a more efficient approach, focusing on lightweight models that maintain high accuracy while minimizing resource consumption, which is an essential requirement for on-satellite processing.
To this end, we propose \textsc{FlickCD}, which means quick flick then get great results, pushing the boundaries of the performance-resource trade-off. \textsc{FlickCD} introduces an Enhanced Difference Module (EDM) to amplify critical feature differences between temporal phases while suppressing irrelevant variations such as lighting and weather changes, thereby reducing computational costs in the subsequent change decoder. Additionally, the \textsc{FlickCD} decoder incorporates Local-Global Fusion Blocks, leveraging Shifted Window Self-Attention (SWSA) and Efficient Global Self-Attention (EGSA) to effectively capture semantic information at multiple scales, preserving both coarse- and fine-grained changes. Extensive experiments on four benchmark datasets demonstrate that \textsc{FlickCD} reduces computational and storage overheads by more than an order of magnitude while achieving state-of-the-art (SOTA) performance or incurring only a minor (<1\% F1) accuracy trade-off.  The implementation code is publicly available at \url{https://github.com/xulsh8/FlickCD}.
\end{abstract}

% keywords can be removed

\section{Introduction}
With the rapid advancement of remote sensing technologies, it has become possible to acquire an increasing number of high-resolution remote sensing images. These abundant and high-quality data sources have greatly enhanced the development of various remote sensing applications, among which change detection (CD) has particularly benefited as a key task. The target of CD is to identify the changes in the same region by analyzing multi-temporal remote sensing images acquired at different times. This technique can be applied to various fields, such as disaster damage assessment \cite{chen2022dual, zheng2021building}, urban planning \cite{guo2021deep, lou2024integrating}, agricultural management \cite{huang2018agricultural} and environment monitoring \cite{li2020review}. The CD techniques make it possible to gain a more comprehensive understanding of the regional differences, enabling the prediction of future changes and the formulation of corresponding measures.

With the growing success of deep learning in various computer vision tasks, it has also emerged as a powerful tool for CD. However, deep learning-based CD methods face two major challenges. The first challenge is how to better balance model performance and resource requirements. To achieve better feature extraction capabilities, change detection models often increase their depth or width a lot. For example, RCTNet \cite{gao2024relating} increases depth by stacking convolutional layers to fuse multi-scale feature maps, while ChangeMamba \cite{chen2024changemamba} expands the receptive field using state-space models to capture global semantic information. However, these operations require a substantial number of parameters and extensive computational resources, making it challenging to deploy them on resource-constrained devices, such as drones and satellites, which are commonly used for acquiring remote sensing images. Currently, several studies have also been conducted on lightweight change detection models \cite{lei2023ultralightweight, codegoni2023tinycd}. Nevertheless, achieving a lightweight design often results in performance degradation and leads to a reduced ability to extract difference information.

The second challenge is how to ignore irrelevant differences while focusing on capturing significant changes. Irrelevant differences can arise due to various factors, including lighting variations, atmospheric conditions, and seasonal changes. To extract the critical changed information between two temporal images, various change detection models employ different approaches. Most models obtain difference information by subtracting the feature maps and taking the absolute value. \cite{chen2021remote, lei2023ultralightweight, gao2024relating}. Some models merge the two temporal feature maps along the channel dimension and use convolution operations to extract difference information \cite{bandara2022transformer, codegoni2023tinycd}, while others perform layer-wise or multi-level feature fusion to achieve this \cite{chen2024changemamba, chen2024multi}. However, each of these methods has its own limitations: some of them fail to filter out irrelevant changes, making it difficult for the model to distinguish between insignificant differences and critical changes. Others require substantial computational resources and a large number of parameters to obtain difference information, thereby increasing the overall resource demand of the model.

\begin{figure}[t]
  \centering
  \includegraphics[width=0.5\linewidth]{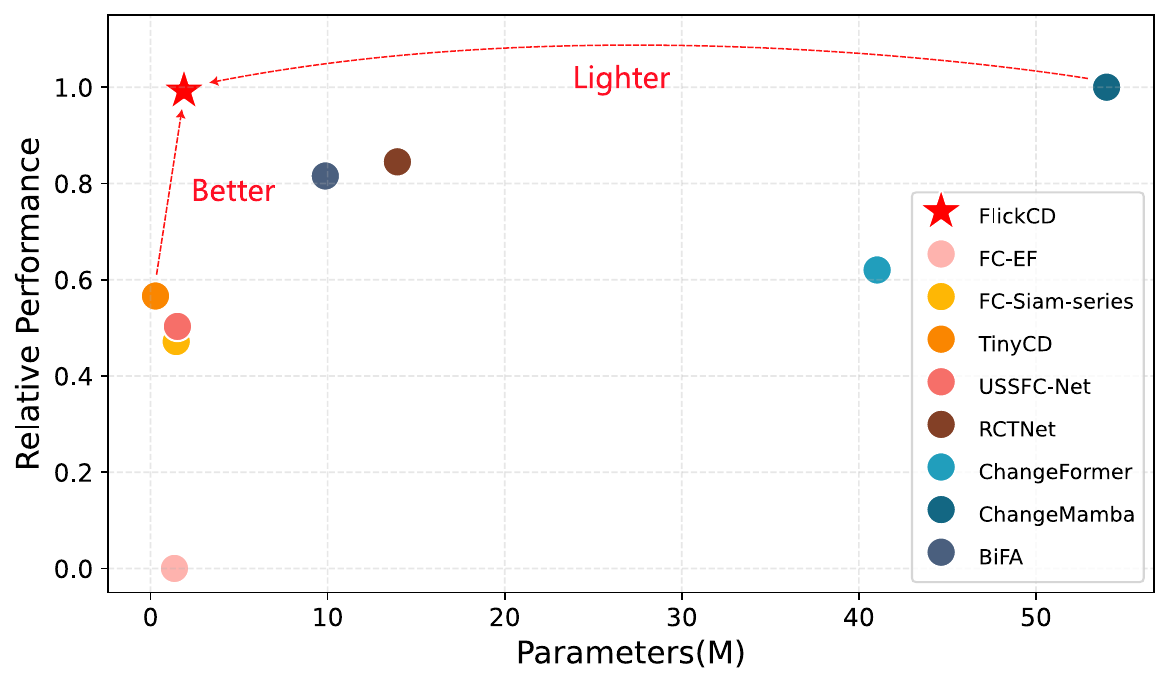}
  \caption{Comparison of parameters and relative performance between FlickCD (Ours) and others. The relative performance is calculated by normalizing the F1 scores across the four datasets, with the lowest value set to 0 and the highest value set to 1.}
  \label{model_comparison_scatter_plot}
\end{figure}

To address the aforementioned challenges, we propose \textsc{FlickCD}, a lightweight model that achieves an excellent balance between resource efficiency and performance. As shown in Fig. \ref{model_comparison_scatter_plot}, we select several lightweight models and state-of-the-art models for comparison. It can be observed that while maintaining a lightweight design, \textsc{FlickCD} achieves an average performance close to that of state-of-the-art models.

\textsc{FlickCD} adopts an encoder-decoder architecture, with RepViT \cite{wang2024repvit} chosen as its encoder. This encoder has the advantages of low latency and a small number of parameters. Unlike most change detection models, we utilize only the feature maps extracted from the first three stages of the encoder, as we observed that incorporating higher-level features does not lead to significant performance improvement, while substantially increasing the number of parameters and computational cost. After extracting feature maps from two temporal images using a shared-weight encoder, we employ the Enhanced Difference Module (EDM) to capture the difference information between them. The lightweight EDM can enhance the difference features while suppressing the irrelevant noise, thereby helping the decoder more accurately extract the changed features of interest. Then, we design the decoder using a bottom-up, layer-by-layer fusion approach. Before fusion, we incorporate a Local-Global Fusion Block (LGFB) to further integrate features. LGFB primarily utilizes attention mechanisms and consists of two components: Sliding-Window Self-Attention (SWSA) and Efficient Global Self-Attention (EGSA). SWSA uses window-based attention to constrain focus to local regions, which helps the model better capture spatial relationships between nearby pixels and improves the accuracy of local object recognition. Subsequently, EGSA is employed to consider the relationship between pixels across the global scope, which is particularly critical when the changed regions consist of a lot of similar small areas. After multiscale extraction and fusion, the process yields a high-quality binary change map as the final result.

Through extensive experiments, we validate the effectiveness of the modules in \textsc{FlickCD} and demonstrate that \textsc{FlickCD} achieves a better trade-off and higher efficiency compared to other models. Our contributions can be summarized as follows:
\begin{itemize}
\item We develop the lightweight change detection model \textsc{FlickCD} and design an efficient lightweight decoder capable of extracting both local and global semantic information effectively.
\item We propose the Enhanced Difference Module (EDM), which filters out irrelevant change noise while preserving critical difference information, thereby helping the model accurately capture change information.
\item We conduct experiments to validate the efficiency of \textsc{FlickCD} among current CD models. It achieves low parameter count and computation cost while maintaining high performance, and even achieves state-of-the-art on some datasets.
\end{itemize}

\section{Related Work}
\subsection{Deep Learning-based Change Detection}
With the remarkable success and high performance of deep learning techniques, the change detection (CD) community has increasingly adopted them to address the CD task. One of the earliest CNN-based approaches is the FC series \cite{daudt2018fully}, which introduced two fundamental strategies for handling dual-temporal inputs: (1) early fusion, where the two input images are first concatenated and then fed into the network; and (2) Siamese architectures, which process the two images with shared-weight encoders followed by a difference computation. These two strategies have laid the foundation for most subsequent CD models. After that, the emergence of Transformer architectures, which leverage the attention mechanisms, has brought new possibilities to CD. For instance, BIT \cite{chen2021remote} adopts a CNN backbone for encoding and decoding, while employing a Transformer to capture differences between the input images. ChangeFormer \cite{bandara2022transformer} further advances this direction by directly applying a shared-weight Transformer to handle the CD task. In addition to the conventional global self-attention, various attention mechanisms tailored for CD have also been explored. These include spatial and channel attention \cite{fang2021snunet, zhang2020deeply, peng2020optical}, as well as the cross-attention mechanism \cite{chen2023land}, which enhance the effectiveness and robustness of CD models.

\subsection{Lightweight Strategies in Deep Learning}
Striking a perfect balance between model efficiency and performance is always a critical research direction. To enable the development of deep learning models on resource-constrained devices, a variety of lightweight strategies have been proposed across different model architectures. In CNN-based strategies, Depth-wise Separable Convolution \cite{howard2017mobilenets} extracts features by decoupling the spatial and channel dimensions. Group Convolution \cite{zhang2018shufflenet} divides channels into groups and performs convolution operations within each group independently. More recently, FasterNet \cite{chen2023run} introduces Partial Convolution, which computes only a subset of channels in the feature maps. In Transformer-based strategies, they are primarily implemented in two ways. First, limit the attention computation scope to local windows or sparse regions \cite{wang2024repvit, liu2021swin, wang2020axial}. Although this strategy may result in information loss, allocating limited parameters to more likely relevant tokens can lead to minimal performance degradation while reducing redundant information. Second, utilize weight-sharing mapping to obtain the attention matrix \cite{liu2021swin, mehta2021mobilevit}. Applying the same linear mapping to correlated regions reduces parameter count and projects them into the same high-dimensional space. There has also been research on lightweight models for CD tasks. USSFC-Net \cite{lei2023ultralightweight} uses dilated convolutions cyclically as the basic module, and constructs the Encoder and Decoder by stacking this module. TinyCD \cite{codegoni2023tinycd} employs the three-stage EfficientNet \cite{tan2019efficientnet} as its backbone and utilizes group convolution to compute difference information. Although these networks have a low number of parameters, they suffer from high latency and lower performance.
\begin{figure*}[t]
  \begin{center}
  \includegraphics[width=0.9\linewidth, keepaspectratio]{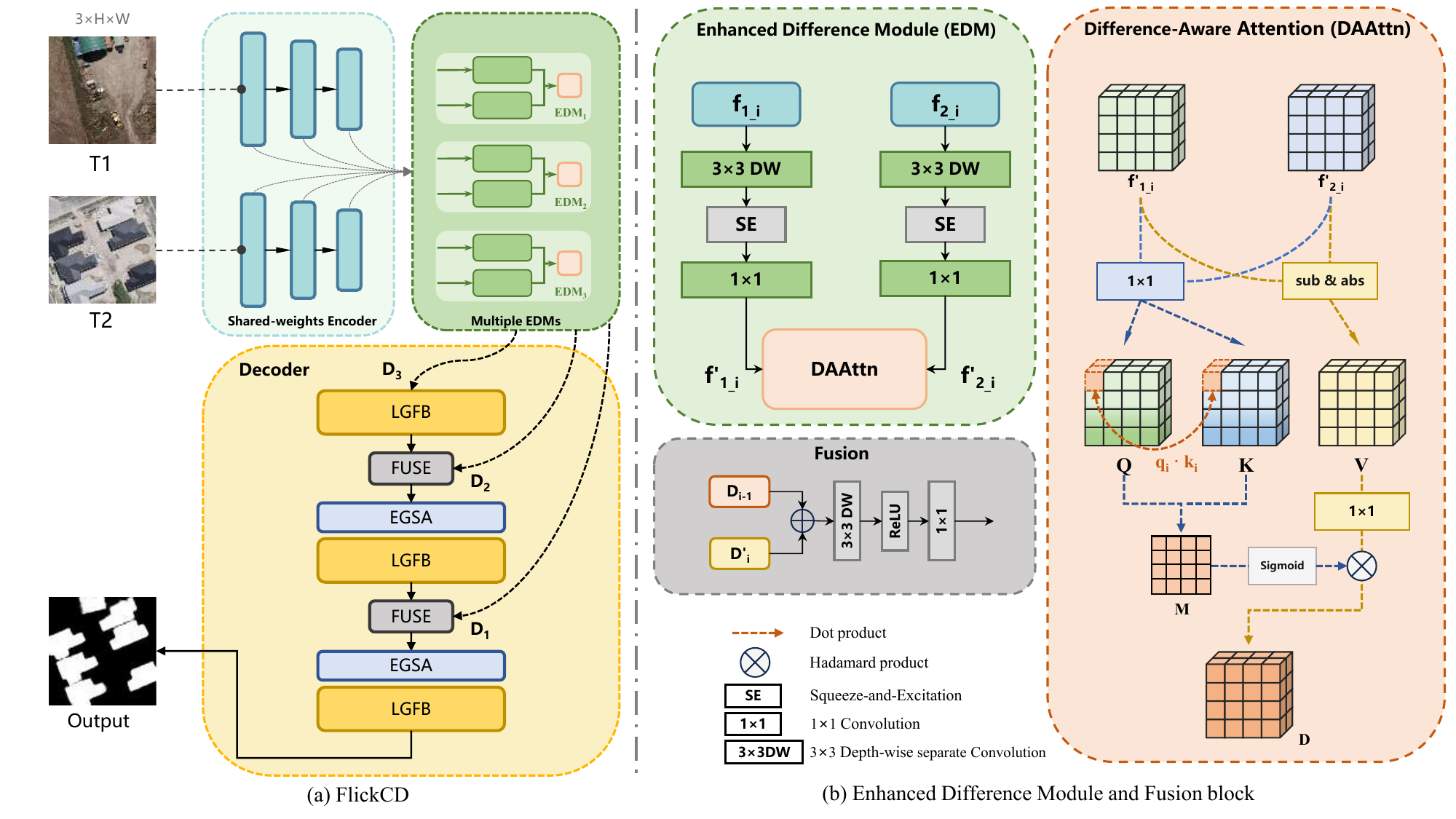}
  \end{center}
  \caption{(a) Overall architecture of the FlickCD model. (b) Structural illustration of the Enhanced Difference Module (EDM) and Fusion module.}
  \label{model}
\end{figure*}

\section{Method}
\subsection{Overview}
The overview of the model structure is shown in the Fig. \ref{model}(a). At first, input the bi-temporal images, $\mathbf{T}_1$ and $\mathbf{T}_2$, both with dimensions $W\times H\times3$, into the weight-sharing encoder. We selected the lightest version of RepViT \cite{wang2024repvit} with pre-trained weights as the encoder. Then, we only use the feature maps from the first three stages of the encoder. On one hand, our experiments demonstrate that the feature map from the fourth stage does not improve the model's performance. On the other hand, removing the fourth stage can reduce the number of parameters and computational complexity. This process can be expressed as:
\[f_{1\_1},f_{1\_2},f_{1\_3} = Encoder(\mathbf{T}_1),\]
\[f_{2\_1},f_{2\_2},f_{2\_3} = Encoder(\mathbf{T}_2).\]
The encoder extracts features hierarchically, where downsampling and channel doubling are applied at each stage. For the feature maps $f_{1\_i}$ and $f_{2\_i},i\in\{1,2,3\}$, the spatial resolution at each stage is $\frac{W}{2^{i+1}} \times \frac{H}{2^{i+1}}$.
Then, we use the Enhanced Difference Module (EDM) to extract the difference feature between $f_1$ and $f_2$ at each stage. Additionally, to facilitate the multi-stage feature fusion in the decoder, the channel dimension of the feature maps from all three stages are unified. We formulate this process as:
\[D_i = EDM_i(f_{1\_i}, f_{2\_i}),i \in \{1,2,3\}.\]
After that, we propose a bottom-up lightweight decoder, which mainly consists of the Local-Global Fusion Block (LGFB) and additional Efficient Global Self-attention (EGSA). Through layer-by-layer fusion and upsampling, it generates the final binary result map.

\subsection{Enhanced Difference Module}
How to extract the difference information between two temporal images effectively is one of the main challenges of the CD task. A common approach for extracting the difference is to directly subtract the two feature maps and take the absolute value. However, since this approach computes pixel-wise differences with equal weights across all pixels, differences caused by irrelevant factors are preserved alongside meaningful change information and passed to the decoder. This makes it difficult for the decoder to accurately identify critical differences, leading to a decrease in accuracy. To address this issue, we propose the Enhanced Difference Module (EDM), which dynamically assigns weights to the pixel-wise differences. This enables the model to filter out irrelevant differences and enhance the identification of critical change information.

The detailed architecture of the EDM is illustrated in Fig. \ref{model}(b). For a given stage $EDM_i$, the inputs are the corresponding feature maps $f_{1\_i}$ and $f_{2\_i}$. Since the magnitude of pixel-wise differences should take local contextual information into account, we first apply a depthwise separable convolution to enhance local feature aggregation. This is followed by a Squeeze-and-Excitation block and a $1\times1$ convolution layer, which are used to perform dynamic channel-wise weighting and adjust the number of output channels accordingly. 

After the above operations, we obtain the feature maps $f'_{1\_i}$ and $f'_{2\_i}$, which are then fed into the Difference-Aware Attention module to generate the difference feature map $D_i$. This module is inspired by the core concept of attention mechanisms: measuring the similarity between vectors via dot products to assign corresponding token weights. Specifically, we first apply identical linear projection to $f'_{1\_i}$ and $f'_{2\_i}$, mapping them into a shared high-dimensional space and obtain the query $\textbf{Q}$ and the key $\textbf{K}$. Next, we compute the dot product between corresponding vectors in $\textbf{Q}$ and $\textbf{K}$ to derive the initial attention mask $\textbf{M}$. Unlike standard attention mechanisms, our goal is to emphasize the features that exhibit greater differences. To this end, we invert the attention mask $\textbf{M}$, such that vector pairs with larger dot products (i.e., similar directions) receive lower weights, while those with smaller dot products (i.e., more distinct features) are assigned higher weights. This inversion encourages the model to focus more on significantly different features. Finally, we apply a Sigmoid function to normalize the inverted scores, resulting in the final attention mask $\textbf{M}'$. To mitigate the risk of vanishing gradients caused by excessively large dot product values, a scaling factor $\sqrt{d_k}$ (where $d_k$ represents the vector dimension) is applied during the dot product computation. This process can be expressed as:
\[\textbf{Q} = Conv_{1\times1}(f'_{1\_i}), \textbf{K} = Conv_{1\times1}(f'_{2\_i}),\]
\[\textbf{M} \supset m_{ij} = \frac{q_{ij} \cdot k_{ij}}{\sqrt{d_k}}, i\in\{1,2,\dots,W\}; j\in\{1,2,\dots,H\},\]
\[\textbf{M}' = Sigmoid(-\textbf{M}).\]
Finally, we compute the value matrix $\textbf{V}$ by taking the element-wise absolute difference between the feature maps $f'_{1\_i}$ and $f'_{2\_i}$. The final output $D_i$ is then obtained by the Hadamard product of the attention mask $\textbf{M}'$ and value matrix $\textbf{V}$.
\[\textbf{V} = Conv'_{1\times1}(|f'_{1\_i} - f'_{2\_i}|),\]
\[D_i = \textbf{M}' \odot \textbf{V}.\]

\begin{figure*}[t]
  \begin{center}
  \includegraphics[width=0.8\linewidth, keepaspectratio]{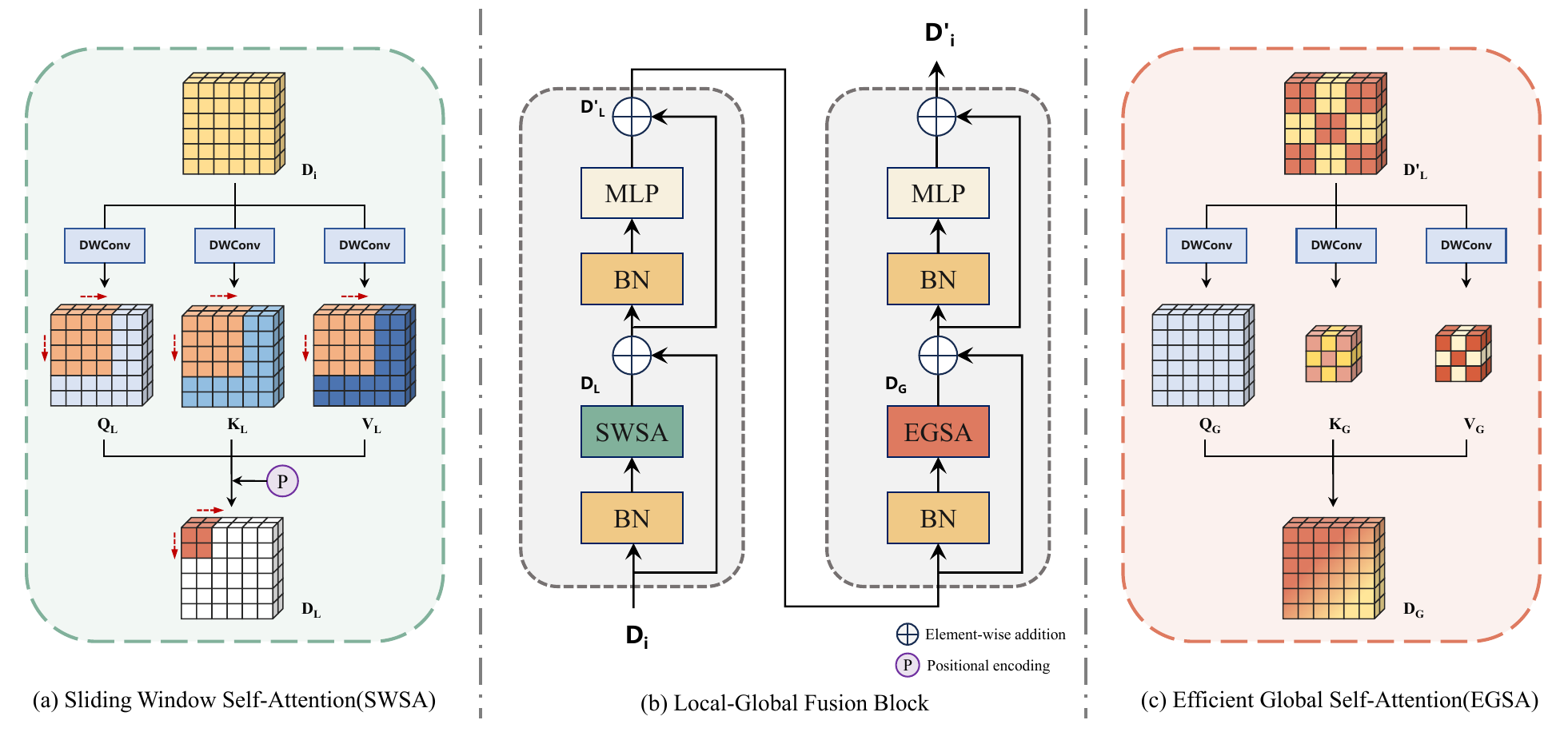}
  \end{center}
  \caption{(a)Structural diagram of the Sliding Window Self-Attention (SWSA) module. (b)Detailed architecture of the Local-Global Fusion Block (LGFB). (c)Structural diagram of the Efficient Global Self-Attention (EGSA) module.}
  \label{LGFB}
\end{figure*}

\subsection{Local-Global Fusion Block}
After obtaining the difference map $D_i$ between the two feature maps, it is essential to further capture the relational information among the changed regions. To this end, we extract relationship features at both local and global scales, motivated by the following observations: 
\begin{itemize}
\item On the local scale, when a specific region is surrounded by other changed areas, it is highly likely that this region has also undergone change. Modeling such local relationships helps mitigate the presence of holes or fragmentation within detected changed regions.
\item On the global scale, by leveraging global contextual information, the model can enhance overall feature representations, thereby increasing the discriminability between changed and unchanged regions and facilitating more accurate delineation of change areas.
\end{itemize}
To achieve the aforementioned objectives, we propose the Local-Global Fusion Block (LGFB). This module consists of two key components: Sliding Window Self-Attention (SWSA) and Efficient Global Self-Attention (EGSA). The former is designed to capture local relational information among pixels within spatially adjacent regions, while the latter focuses on modeling long-range dependencies and enhancing feature representation. The detailed architecture of LGFB is illustrated in Fig. \ref{LGFB}. Structurally, LGFB is a Transformer-like module. However, traditional self-attention mechanisms are often computationally and parameter-intensive. To address this, we adopt a more efficient design, which is described in detail below.

\textbf{Token Mixer \& Channel Mixer.} Inspired by the RepViT \cite{wang2024repvit}, we decompose the attention computation into two stages: a token mixer and a channel mixer, where the former is responsible for aggregating spatial information and the latter focuses on inter-channel interaction. Specifically, in both the SWSA and EGSA modules, we adopt Depth-wise Separable Convolution (DWConv) for linear projections. The projected matrices are then fed into the attention mechanism to efficiently exchange spatial information, acting as a lightweight token mixer. This process can be formally expressed as:
\[Q,K,V = DWConv_1(I), DWConv_2(I), DWConv_3(I),\]
\[O = Sigmoid(\frac{QK^T}{\sqrt{d}})V + I.\]
Here, $I$ represents the input to the SWSA and EGSA modules. Subsequently, a multi-layer perceptron (MLP) composed of two $1\times1$ convolutional layers is employed to perform information exchange along the channel dimension, serving as the channel mixer. This process can be formally expressed as:
\[O' = MLP(BN(O)) + O.\]

\textbf{Sliding Window Self-Attention.}To enable the model to focus on local pixel-level relationships, we constrain the attention computation within a sliding window. A stride parameter $s$ is introduced to control how the window moves across the entire feature map, producing the final attention output. Both the window size $w$ and the stride $s$ are treated as hyperparameters. When the window size is equal to the stride ($w=s$), the feature map is partitioned into non-overlapping patches, and attention is computed independently within each patch. In contrast, when the window size is larger than the stride ($w>s$), only a portion of the computed attention within each window is retained as a patch, resulting in overlapping receptive fields where the patch size is $s\times s$. In scenarios where changed regions exhibit complex shapes, using a larger window with overlapping receptive fields enhances spatial continuity across neighboring regions. Conversely, for regions with low shape complexity, such as regular building changes, setting the window size equal to the stride is sufficient to capture local patterns. It avoids unnecessary contextual redundancy while reducing additional computational overhead.

\textbf{Efficient Global Self-Attention.}To capture global relationships among changed regions and further mitigate the spatial discontinuity introduced by patch-wise processing in SWSA, we introduce Efficient Global Self-Attention (EGSA) following SWSA. In EGSA, we first apply linear projections to the input $D_L$ to obtain the query $Q_G$, key $K_G$, and value $V_G$ matrices. The $Q_G$ matrix retains the same spatial resolution as the input to preserve detailed spatial information, while the $K_G$ and $V_G$ matrices are downsampled to a lower resolution. This downsampling serves to compress global information, effectively reducing computational cost. Specifically, the $K_G$ and $V_G$ matrices are obtained via convolution operations, where both the kernel size and stride are set to match the patch size used in SWSA. This design ensures that each token in $K_G$ and $V_G$ corresponds to a patch from the input $D_L$. We then compute the attention weights between each pixel in $Q_G$ and all patch-level tokens in $K_G$, enabling the model to aggregate long-range contextual information. The final output feature map is generated by applying the computed attention weights to the $V_G$ matrix. 

\textbf{Hierarchical Feature Integration.}The lightweight decoder takes the multi-scale difference features $D_i, i\in\{1,2,3\}$ generated from three hierarchical stages, as input. Among them, higher-level difference maps have lower spatial resolution but capture richer semantic information about changed regions, whereas lower-level maps retain finer spatial details, such as object boundaries, but are less reliable in distinguishing actual changes. To leverage the complementary advantages of different levels, we adopt a progressive fusion strategy. As illustrated in Fig. \ref{model}(a), each higher-level difference map is processed by the LGFB, and then fused with the corresponding lower-level difference map via element-wise addition. A Depth-wise Separable Convolution is then applied to refine the fused features, achieving a local adjustment. However, this local fusion alone is not sufficient to capture long-range dependencies across spatially distant regions. Therefore, we further incorporate EGSA after the local fusion to perform global refinement and enhance the fusion quality.

\begin{figure*}[ht]
  \begin{center}
  \includegraphics[width=0.8\linewidth, keepaspectratio]{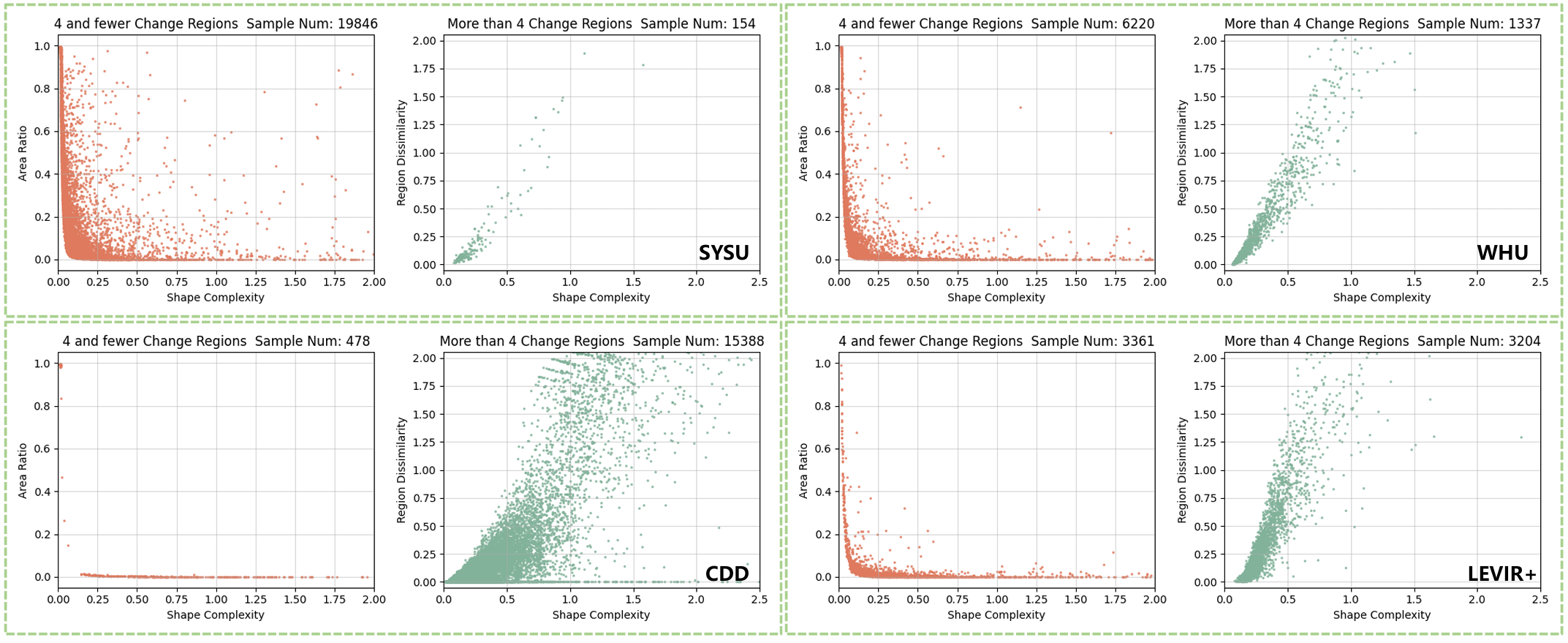}
  \end{center}
  \caption{Changed region characteristics in four datasets. For each dataset, samples are divided by the number of changed regions ($\leq$4 on the left, $>$4 on the right). The x-axis denotes shape complexity (perimeter/area); the y-axis shows area ratio for the left plots and region similarity (variance of shape complexity) for the right plots.}
  \label{Dataset_Eval}
\end{figure*}

\section{Experiment and Discussion}
\subsection{Datasets and Experiment Settings}
To validate the effectiveness of \textsc{FlickCD}, we conduct experiments on four widely-used change detection datasets: SYSU \cite{shi2021deeply}, WHU \cite{ji2018fully}, LEVIR+ \cite{chen2020spatial}, and CDD \cite{lebedev2018change}. To better interpret the experimental results, we perform a statistical analysis on the characteristics of changed regions within each dataset, as shown in Fig. \ref{Dataset_Eval}.
We categorize the samples in each dataset into two groups based on the number of changed regions: few-region samples and many-region samples, using a threshold of 4 changed regions. For few-region samples, we analyze the area proportion of changed regions and their shape complexity. For many-region samples, we focus on the inter-region dissimilarity and shape complexity. The analysis reveals that:
\begin{itemize}
    \item SYSU primarily consists of samples with a small number of changed regions, where most changes occupy a large area and exhibit relatively high shape complexity. 
    \item CDD mainly consists of samples with numerous changed regions, which show large variation in similarity between regions and exhibit complex shapes.
    \item WHU and LEVIR+ focus mainly on building-related changes, leading to a large number of multi-region samples. In these datasets, the changed regions tend to exhibit high inter-region similarity and low shape complexity, reflecting the relatively regular structure of architectural changes.
\end{itemize}

\textbf{Experiment Settings.}
To accommodate the varying characteristics of changed regions across datasets, we configure the hyperparameters of the Sliding-Window Self-Attention (SWSA) module, namely the window size $w$ and stride $s$ individually for each dataset. Given that higher-level feature maps have lower spatial resolution, we assign smaller window sizes at deeper layers accordingly.For the SYSU and CDD datasets, the changed regions tend to exhibit higher shape complexity, requiring a larger receptive field to improve the accuracy of feature extraction. Accordingly, we set the window sizes to 8,8,16 with corresponding strides of 4,4,8 for these datasets. In contrast, the changed regions in WHU and LEVIR+ are generally more regular and structurally simple. For these datasets, we adopt smaller window sizes and strides configuration to reduce computation. Specifically, we set the parameters to 4,4,8 for WHU and 4,8,8 for LEVIR+. For model training, the batch size is set to 32. The learning rate is set to 0.0005 for all datasets except WHU, which uses a lower learning rate of 0.0002. We train the model for 100 epochs on the SYSU and WHU datasets, 200 epochs on LEVIR+, and 250 epochs on CDD.

\begin{table*}[ht]
    \centering
    \renewcommand{\arraystretch}{0.9}
    \setlength{\tabcolsep}{12pt}
    \caption{Comparative Analysis of model complexity and performance on SYSU and WHU datasets. \textbf{Bold} indicates the best result, and \underline{underline} denotes the second-best. All values are presented as percentages(\%).}
    \resizebox{\linewidth}{!}{
    \begin{tabular}{l|cc|ccccc|ccccc}
        \toprule
        \multirow{2}{*}{Methods} & \multirow{2}{*}{Param.} & \multirow{2}{*}{FLOPs(G)} & \multicolumn{5}{c|}{SYSU} & \multicolumn{5}{c}{WHU} \\
        & & & Rec. & Pre. & OA & \textbf{F1} & \textbf{IoU} & Rec. & Pre. & OA & \textbf{F1} & \textbf{IoU} \\
        \midrule
        FC-EF\cite{daudt2018fully} & 1.35M & 6.23 & 77.92 & 78.94 & 89.89 & 78.43 & 64.51 & 73.10 & 87.89 & 99.10 & 79.81 & 66.41 \\
        FC-Siam-Diff\cite{daudt2018fully} & 1.35M & 8.51 & 79.05 & 80.03 & 90.41 & 79.54 & 66.03 & 85.06 & 90.02 & 99.41 & 87.47 & 77.73 \\
        FC-Siam-Conc\cite{daudt2018fully} & 1.55M & 9.72 & 79.25 & 79.48 & 90.28 & 79.36 & 65.79 & 86.75 & 90.82 & 99.46 & 88.74 & 79.76 \\
        TinyCD\cite{codegoni2023tinycd} & 285.12K & 2.81 & 80.08 & 80.76 & 90.80 & 80.42 & 67.25 & 87.30 & 91.93 & 99.50 & 89.55 & 81.09 \\
        USSFC-Net\cite{lei2023ultralightweight} & 1.52M & 6.32 & 79.23 & 78.18 & 89.89 & 78.70 & 64.88 & 90.15 & 92.75 & 99.59 & 91.43 & 84.21 \\
        \midrule
        ChangeFormer\cite{bandara2022transformer}& 41.03M & 276.81 & 75.61 & 82.61 & 90.49 & 78.95 & 65.23 & 90.44 & 94.09 & 99.62 & 92.23 & 85.58 \\
        RCTNet\cite{gao2024relating}& 13.94M & 23.99 & 78.64 & 84.79 & 91.64 & 81.60 & 68.92 & 91.58 & 95.09 & 99.68 & 93.30 & 87.44 \\
        ChangeMamba\cite{chen2024changemamba}& 53.99M & 61.69 & 78.61 & \textbf{88.27} & 92.49 & 83.16 & 71.17 & \textbf{93.58} & \underline{95.57} & \underline{99.74} & \underline{94.56} & \underline{89.69} \\
        BiFA\cite{zhang2024bifa}& 9.87M & 105.75 & \textbf{80.43} & 87.52 & \underline{92.68} & \underline{83.83} & \underline{72.16} & 91.37 & 95.50 & 99.68 & 93.39 & 87.60 \\
        \midrule
        FlickCD(Ours) & 1.89M & 4.16 &\underline{80.28} & \underline{88.03} & \textbf{92.77} & \textbf{83.97} & \textbf{72.38} & \underline{93.18} & \textbf{96.51} & \textbf{99.75} & \textbf{94.81} & \textbf{90.14} \\
        \bottomrule
    \end{tabular}
    }
    \label{SYSU&WHU}
\end{table*}

\begin{table*}[ht]
    \centering
    \renewcommand{\arraystretch}{0.9}
    \setlength{\tabcolsep}{12pt} 
    \caption{The model complexity and performance comparison of different models on CDD and LEVIR+ datasets.}
    \resizebox{\linewidth}{!}{
    \begin{tabular}{l|cc|ccccc|ccccc}
        \toprule
        \multirow{2}{*}{Method} & \multirow{2}{*}{Param.} & \multirow{2}{*}{FLOPs(G)} & \multicolumn{5}{c|}{CDD} & \multicolumn{5}{c}{LEVIR+} \\
        & & &Rec. & Pre. & OA & \textbf{F1} & \textbf{IoU} & Rec. & Pre. & OA & \textbf{F1} & \textbf{IoU} \\
        \midrule
        FC-EF\cite{daudt2018fully} & 1.35M & 6.23 & 91.46 & 93.94 & 98.13 & 92.68 & 86.37 & 74.42 & 74.04 & 97.89 & 74.23 & 59.02 \\
        FC-Siam-Diff\cite{daudt2018fully} & 1.35M & 8.51 & 94.44 & 95.91 & 98.76 & 95.17 & 90.79 & 80.05 & 80.28 & 98.39 & 80.17 & 66.90 \\
        FC-Siam-Conc\cite{daudt2018fully} & 1.55M & 9.72 & 93.04 & 95.20 & 98.49 & 94.11 & 88.87 & 80.92 & 81.93 & 98.50 & 81.42 & 68.67 \\
        TinyCD\cite{codegoni2023tinycd} & 285.12K & 2.81 & 93.40 & 94.73 & 98.48 & 94.06 & 88.79 & 82.23 & 82.86 & 98.58 & 82.54 & 70.28 \\
        USSFC-Net\cite{lei2023ultralightweight} & 1.52M & 6.32 & 96.57 & 93.99 & 98.87 & 95.26 & 90.95 & 77.19 & 80.46 & 98.31 & 78.79 & 65.00 \\
        \midrule
        ChangeFormer\cite{bandara2022transformer} & 41.03M & 276.81 & 94.51 & 94.82 & 98.74 & 94.67 & 89.87 & 81.45 & 84.14 & 98.61 & 82.77 & 70.61 \\
        RCTNet\cite{gao2024relating}& 13.94M & 23.99 & 96.99 & 97.80 & 99.33 & 97.39 & 94.92 & 84.58 & 85.09 & 98.77 & 84.83 & 73.66 \\
        ChangeMamba\cite{chen2024changemamba}& 53.99M & 61.69 & \textbf{98.21} & \underline{97.66} & \textbf{99.51} & \textbf{97.93} & \textbf{95.95} & \textbf{86.99} & \underline{87.70} & \textbf{98.97} & \textbf{87.34} & \textbf{77.53} \\
        BiFA\cite{zhang2024bifa}& 9.87M & 105.75 & 95.63 & 95.64 & 98.97 & 95.63 & 91.63 & 84.44 & 81.93 & 98.61 & 83.16 & 71.18 \\
        \midrule
        FlickCD(Ours)& 1.89M & 4.16 & \underline{97.09} & \textbf{98.18} & \underline{99.39} & \underline{97.63} & \underline{95.37} & \underline{84.59} & \textbf{88.09} & \underline{98.91} & \underline{86.30} & \underline{75.90} \\
        \bottomrule
    \end{tabular}
    }
    \label{CDD&LEVIR+}
\end{table*}

\subsection{Overall Comparison}
To demonstrate the trade-off advantage of \textsc{FlickCD} in CD tasks, we compare it against both lightweight and high-performance CD models. The lightweight models include the FC series (FC-EF, FC-Siam-Diff, FC-Siam-Conc) \cite{daudt2018fully}, TinyCD \cite{codegoni2023tinycd}, and USSFC-Net \cite{lei2023ultralightweight}. The high-performance models include RCTNet \cite{gao2024relating}, ChangeFormer \cite{bandara2022transformer}, BiFA \cite{zhang2024bifa}, and ChangeMamba \cite{chen2024changemamba}. Through this comparison, we aim to highlight the efficiency advantage of \textsc{FlickCD} over high-performance models, and its performance advantage over lightweight models.

\textbf{Quantitative Results.}
We primarily compare the F1 scores of different models across four datasets, as this metric provides a comprehensive evaluation of model performance. As shown in Table \ref{SYSU&WHU} and \ref{CDD&LEVIR+}, it is evident that models with fewer parameters than \textsc{FlickCD} perform significantly worse across all datasets. Compared with the models that have more parameters, \textsc{FlickCD} achieves the highest performance in SYSU and WHU datasets, and maintains performance on par with the best-performing model ChangeMamba in CDD and LEVIR+ datasets. Although \textsc{FlickCD} does not outperform ChangeMamba in terms of F1 score on the CDD and LEVIR+ datasets, it maintains a comparable performance level and ranks second among the selected models.

\begin{figure}[h]
  \begin{center}
  \includegraphics[width=0.7\linewidth, keepaspectratio]{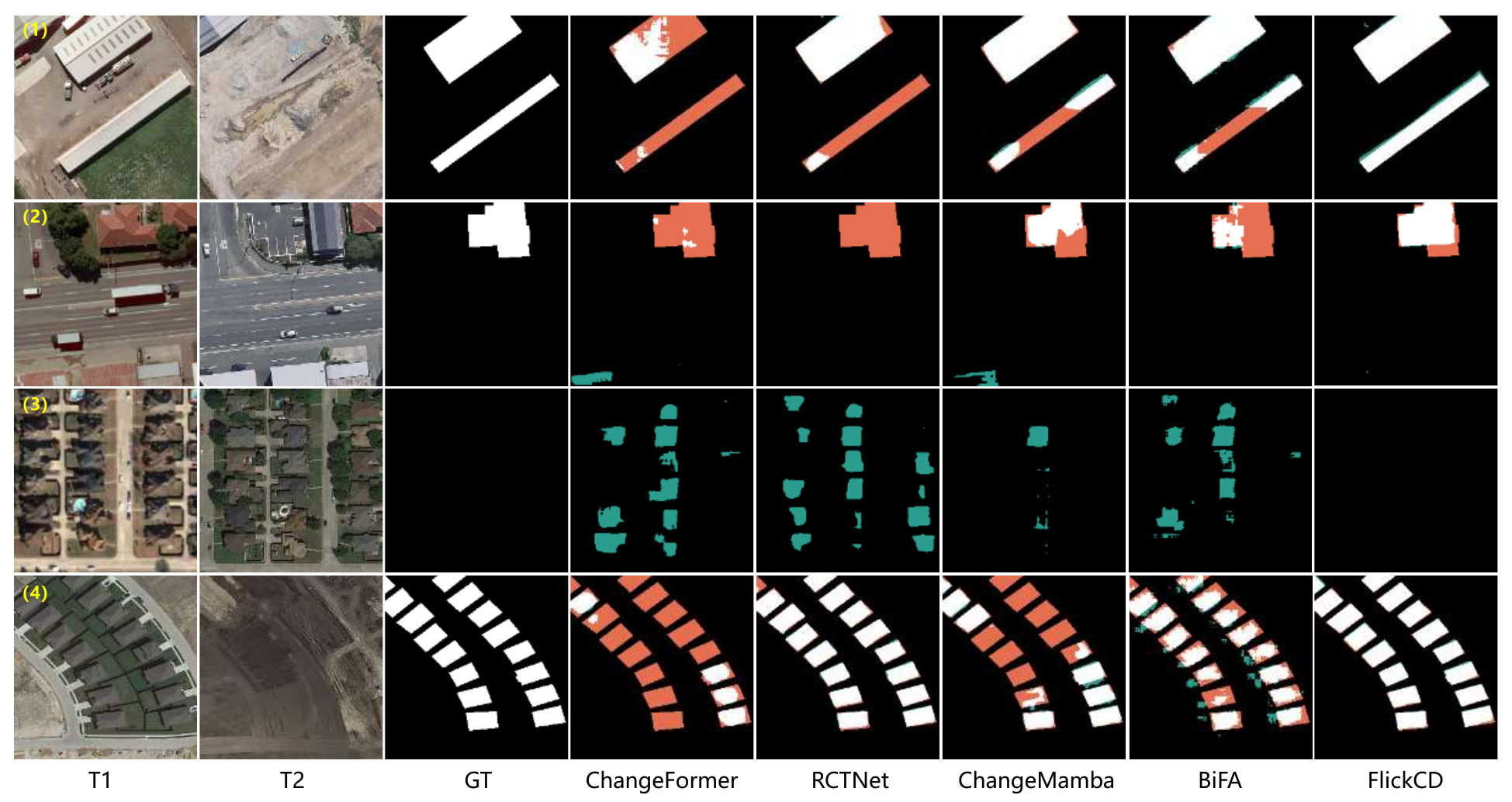}
  \end{center}
  \caption{Visualization results on the WHU and LEVIR+ dataset, where white represents true positive, black represents true negative, green represents false positive, and red represents false negative.}
  \label{WHU&LEVIR_Qual}
\end{figure}

\textbf{Qualitative Results.} 
To gain a more intuitive and detailed understanding of model performance in CD tasks, we select samples from the SYSU, WHU, and LEVIR+ datasets and compare the output of five resource-intensive models with \textsc{FlickCD}. As shown in Fig. \ref{WHU&LEVIR_Qual} and \ref{SYSU_Qual}, \textsc{FlickCD} demonstrates superior recognition performance on challenging changed regions. Specifically, for Fig. \ref{WHU&LEVIR_Qual}(3) and Fig. \ref{SYSU_Qual}, some color variations caused by illumination differences are actually irrelevant changes. However, the comparison models mistakenly classify them as significant changes, while \textsc{FlickCD} identifies and filters out these irrelevant differences. Furthermore, although \textsc{FlickCD} achieves a slightly lower F1-score than ChangeMamba on the LEVIR+ dataset, it shows better extraction performance in cases involving multiple similar changed regions, such as Fig. \ref{WHU&LEVIR_Qual}(4). This visual comparison highlights \textsc{FlickCD}’s stronger ability to distinguish irrelevant differences and more accurate extraction for changed regions.

\textbf{Model Efficiency Analysis}
To compare the computational efficiency and resource requirements of different models, we report the number of parameters, FLOPs, throughput, and inference time for each model. Specifically, the inference time refers to the time required to process a single sample, while the throughput is calculated by dividing the number of samples in one epoch by the total runtime of that epoch. The corresponding results are presented in Table \ref{SYSU&WHU} and \ref{effiTable}. The FC series, TinyCD, and USSFC-Net significantly reduce the parameter count at the cost of considerable performance degradation. However, the FC series and USSFC-Net still have relatively high FLOPs values. Among the high-performance models, although they achieve competitive results across various datasets, their resource demands are substantial. For example, ChangeMamba, which demonstrates a good balance of accuracy across datasets, requires 53.99M parameters and 31.3 ms inference time per sample. In contrast, \textsc{FlickCD} achieves comparable performance with only 1.89M parameters, approximately 3.5\% of ChangeMamba, and a lower inference time of 9.2 ms, which is less than one-third of that of ChangeMamba. In addition, \textsc{FlickCD} achieves a throughput that is 2 to 3 times higher than that of the high-performance models. These results clearly demonstrate the efficiency of \textsc{FlickCD} in terms of parameter count, computation cost, and latency, while maintaining strong performance.

\begin{table}
    \centering
    \renewcommand{\arraystretch}{0.8}
    \caption{Comparison of Throughput and Inference Time across different models. Higher Throughput and Lower Inference Time Indicate Better Performance.}
    \resizebox{0.5\linewidth}{!}{
    \begin{tabular}{l|cc}
        \toprule
        Methods & Throughput(samples/s) & Inference Time (ms) \\
        \midrule
        FC-EF\cite{daudt2018fully} & 402.82 & 1.5\\
        FC-Siam-Diff\cite{daudt2018fully} & 323.10 & 2.1\\
        FC-Siam-Conc\cite{daudt2018fully} & 326.00 & 2.1\\
        TinyCD\cite{codegoni2023tinycd} & 166.81 & 5.0\\
        USSFC-Net\cite{lei2023ultralightweight} & 68.66 & 13.4\\
        \midrule
        ChangeFormer\cite{bandara2022transformer} & 40.42 & 11.4\\
        RCTNet\cite{gao2024relating} & 56.03 & 16.7\\
        ChangeMamba\cite{chen2024changemamba} & 30.82 & 31.3\\
        BiFA\cite{zhang2024bifa} & 32.32 & 19.8\\
        \midrule
        FlickCD (Ours) & \textbf{95.08} & \textbf{9.2}\\
        \bottomrule
    \end{tabular}
    }
    \label{effiTable}
\end{table}

\subsection{Ablation Study}
\begin{figure}[h]
  \centering
  \includegraphics[width=0.7\linewidth, keepaspectratio]{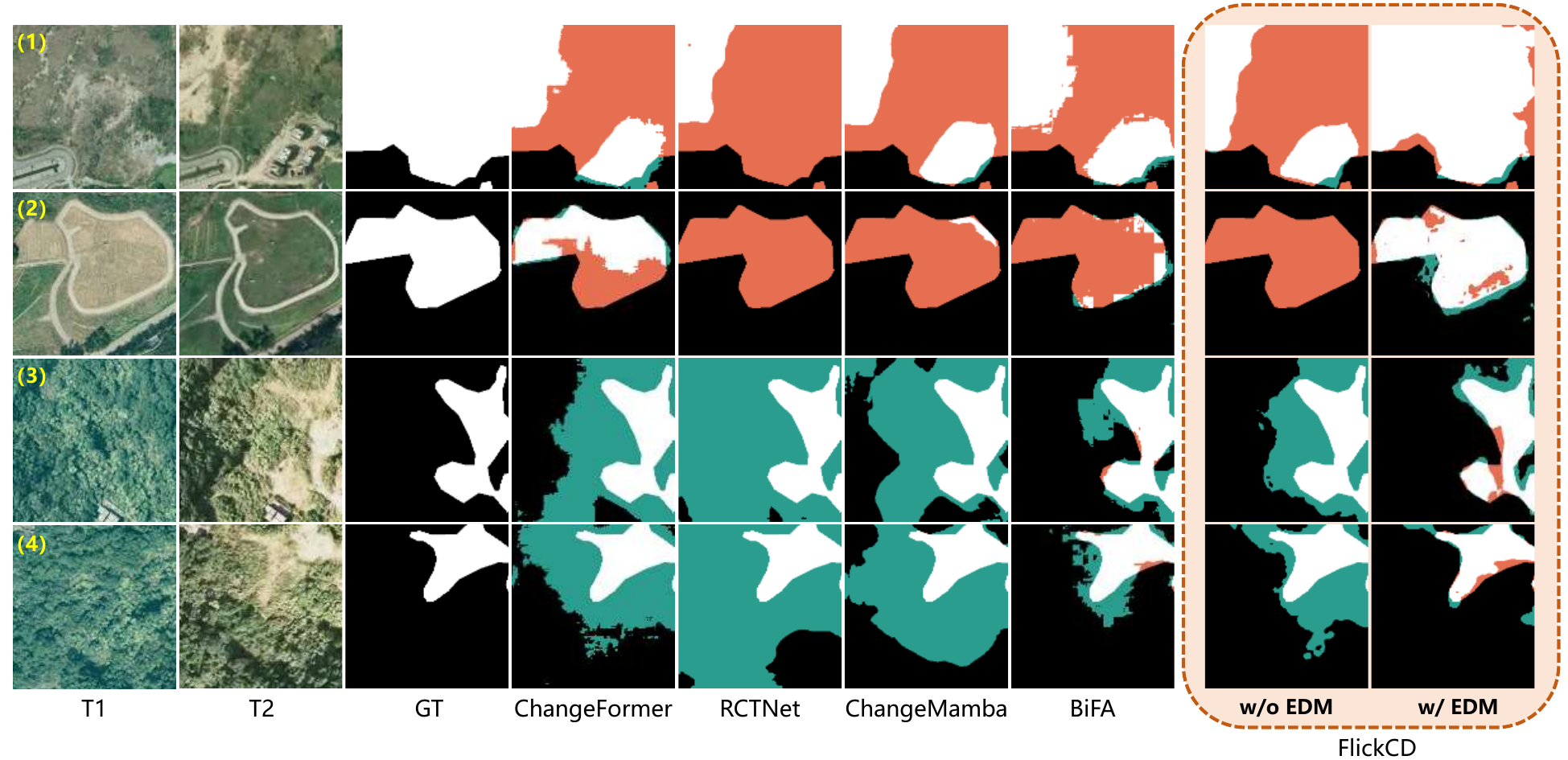}
  \caption{Visualization results on the SYSU dataset, showcasing the outputs of other models, FlickCD with EDM, and FlickCD without EDM.}
  \label{SYSU_Qual}
\end{figure}

\textbf{Effectiveness of EDM}
To validate the effectiveness of the Enhanced Difference Module (EDM) in \textsc{FlickCD}, we first conducted a visual ablation study, with the results shown in Fig. \ref{SYSU_Qual}. On the SYSU dataset, \textsc{FlickCD} exhibits a stronger ability to distinguish between irrelevant and critical changes. To confirm that this capability stems from the EDM, we removed EDM and retrained the model, obtaining the difference features between the two inputs by subtracting their feature maps and taking the absolute value. As illustrated in the results, the model fails to distinguish irrelevant changes from significant ones without EDM, similar to other baseline methods. In addition to the visual comparison, we also performed quantitative ablation experiments across all datasets, as shown in Table \ref{ModuleAbalation}. By integrating the EDM into the base architecture, we observe consistent improvements in F1-scores across all datasets, with the most significant gain of 0.65\% on SYSU. These findings from both visual and quantitative perspectives demonstrate the effectiveness of the EDM in enhancing the model’s discriminative capability.

\begin{table}
    \centering
    \renewcommand{\arraystretch}{0.8} 
    \caption{Ablation Study of Key Modules. All values are presented as percentages(\%).}
    \resizebox{0.5\linewidth}{!}{
    \begin{tabular}{ccc|cccc}
        \toprule
        \multicolumn{3}{c|}{Module} & \multicolumn{4}{c}{F1} \\
        \midrule
        EDM & SWSA & EGSA & SYSU & WHU & CDD & LEVIR+\\
        \midrule
        \textcolor{red}{\ding{55}} & \textcolor{red}{\ding{55}} & \textcolor{red}{\ding{55}} & 80.41 & 93.34 & 96.60 & 85.91 \\
        \ding{51} & \textcolor{red}{\ding{55}} & \textcolor{red}{\ding{55}} & 81.06 & 93.98 & 96.77 & 86.19\\
        \ding{51} & \ding{51} & \textcolor{red}{\ding{55}} & 82.62 & 93.73 & 97.18 & 85.63\\
        \ding{51} & \textcolor{red}{\ding{55}} & \ding{51} & 82.66 & 93.89 & 97.33 & 85.94\\
        \midrule
        \ding{51} & \ding{51} & \ding{51} & \textbf{83.97} & \textbf{94.81} & \textbf{97.63} & \textbf{86.30}\\
        \bottomrule
    \end{tabular}
    }
    \label{ModuleAbalation}
\end{table}

\textbf{Effectiveness of SWSA and EGSA.}
To verify the effectiveness of Sliding-Window Self-Attention (SWSA) and Efficient Global Self-Attention (EGSA) in analyzing difference information, we conducted ablation experiments involving these two modules on all datasets. The results are summarized in Table \ref{ModuleAbalation}, revealing two distinct trends:
\begin{itemize}
\item For the SYSU and CDD datasets, introducing either module individually leads to performance improvements, and combining both results in further gains.
\item In contrast, on the WHU and LEVIR+ datasets, using only one of the modules causes a performance drop, while combining both modules yields a notable performance increase.
\end{itemize}
These findings, when correlated with the dataset characteristics illustrated in Fig. \ref{Dataset_Eval}, provide further insights. Both WHU and LEVIR+ datasets contain a large number of multi-instance changed regions, with high similarity among regions and low shape complexity. In such cases, omitting one attention scale (local or global) leads to a greater loss than the gain achieved by using only one. This suggests that both local and global contextual analyses are essential for effectively capturing meaningful changed information in datasets characterized by numerous, similar, and relatively simple changed regions.

\begin{figure}[h]
  \centering
  \includegraphics[width=0.7\linewidth, keepaspectratio]{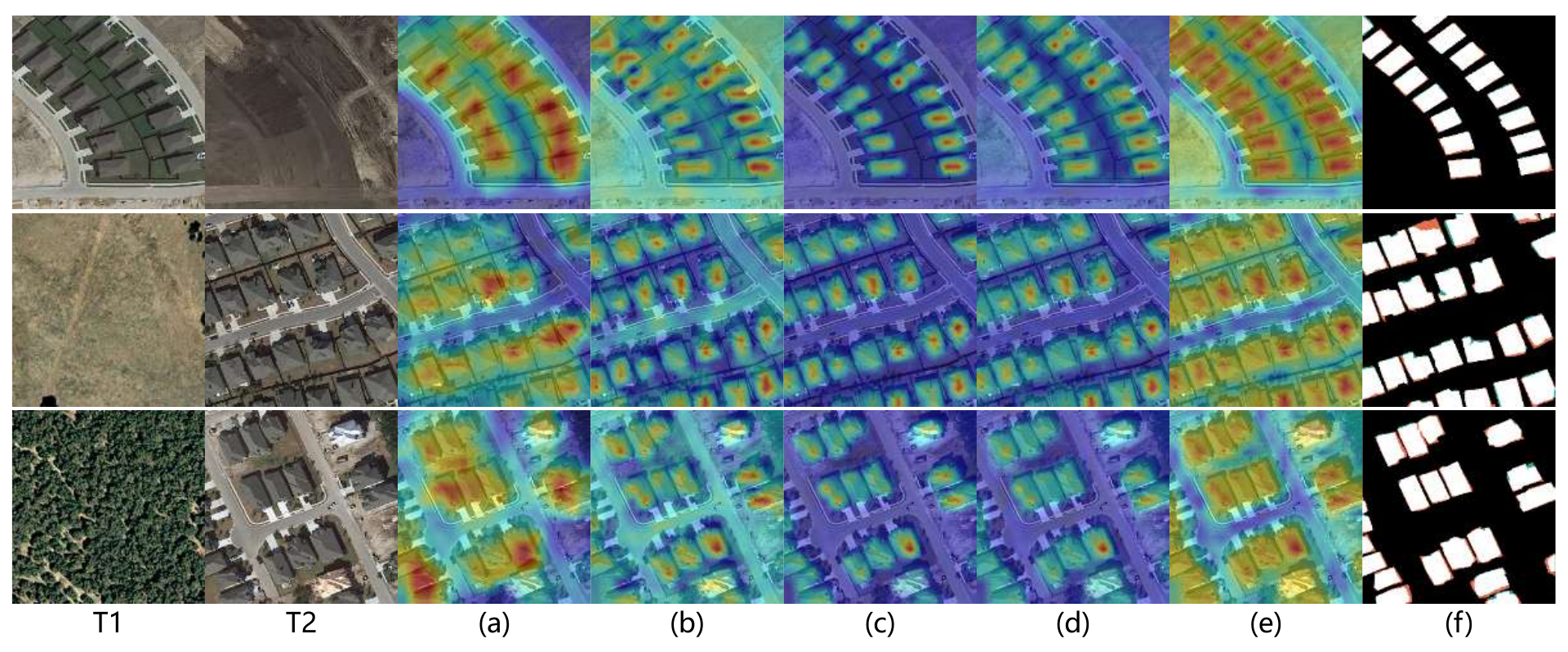}
  \caption{Visualization of intermediate feature maps and attention heatmaps: (a) Difference map before EDM; (b) Attention heatmap of EDM; (c) Feature map after EDM; (d) Feature map after SWSA; (e) Feature map after EGSA; (f) Final result.}
  \label{FeatureMaps}
\end{figure}

\textbf{Visualization of Intermediate Features.}
To further demonstrate the effectiveness of each module, we visualize several intermediate feature maps and attention heatmaps, as shown in Fig. \ref{FeatureMaps}. The difference map before EDM processing fails to accurately detect the changed regions. In contrast, the attention heatmap generated by EDM reallocates the weights of different regions, resulting in a more precise segmentation of the changed areas (Fig. \ref{FeatureMaps}(c)). Subsequently, SWSA refines the delineation of local changed regions, while EGSA further enhances the global difference information. Finally, the model outputs an accurate binary change detection map.

\begin{table}[h]
    \centering
    \renewcommand{\arraystretch}{0.8}
    \caption{Comparison of the parameter reduction and performance impact of two lightweight strategies, w/o dpConv indicates that depthwise separable convolutions are not used in SWSA and EGSA. Four stages refers to using outputs from all four stages of the backbone.}
    \resizebox{0.5\linewidth}{!}{
    \begin{tabular}{c|c|cccc}
        \toprule
        \multirow{2}{*}{Strategy} & \multirow{2}{*}{Param.} & \multicolumn{4}{c}{F1} \\
        & & SYSU & WHU & CDD & LEVIR+ \\
        \midrule
        Base & \textbf{1.89M} & \textbf{83.97} & \textbf{94.81} & \textbf{97.63} & \textbf{86.30}\\
        w/o dpConv & 3.60M & 83.51 & 94.05 & 97.54 & 85.68\\
        Four stages & 3.19M & 82.67 & 94.05 & 96.99 & 85.99\\
        \bottomrule
    \end{tabular}
    }
    \label{LightStrategy}
\end{table}
\textbf{Lightweight Strategy.}
In \textsc{FlickCD}, we investigated two lightweight design strategies that potentially introduce some information loss but significantly reduce the model size: 
\begin{itemize}
\item Employing depth-wise separable convolutions for linear projections in both EGSA and SWSA.
\item Using feature maps from only the first three encoder stages for change detection.
\end{itemize}
To assess the potential trade-offs introduced by these strategies, we conducted ablation experiments, with the results summarized in Table \ref{LightStrategy}. The findings indicate that both strategies reduce the model’s parameter count by nearly half, yet no performance degradation was observed across the four datasets. This suggests that the information loss induced by these strategies is non-critical. Specifically, while depth-wise separable convolutions discard inter-channel interactions, this limitation is compensated by the subsequent MLP, which facilitates cross-channel communication. In addition, the third encoder stage already provides sufficient semantic information for changed region extraction, and incorporating the fourth stage does not further improve performance.

\section{Conclusion}
We propose a lightweight model, \textsc{FlickCD}, which maintains high performance while minimizing resource requirements. The model is primarily composed of convolution operations and attention mechanisms. Specifically, we design the Enhanced Difference Module (EDM) to enhance bi-temporal image difference information while reducing irrelevant noise. Additionally, we introduce two lightweight feature extraction modules, SWSA and EGSA, which effectively extract the local and global features while maintaining efficiency. To evaluate the effectiveness of \textsc{FlickCD}, we conducted extensive experiments on four datasets, analyzing its resource consumption and inference speed, along with ablation studies. The results consistently demonstrate our model's low computational cost and high performance. This research provides new insights and methodologies for deploying change detection models on resource-constrained devices in the future.

\bibliographystyle{unsrt}  
\bibliography{references}

\begin{thebibliography}{10}

\bibitem{chen2022dual}
Hongruixuan Chen, Edoardo Nemni, Sofia Vallecorsa, Xi~Li, Chen Wu, and Lars Bromley.
\newblock Dual-tasks siamese transformer framework for building damage assessment.
\newblock In {\em IGARSS 2022-2022 IEEE International Geoscience and Remote Sensing Symposium}, pages 1600--1603. IEEE, 2022.

\bibitem{zheng2021building}
Zhuo Zheng, Yanfei Zhong, Junjue Wang, Ailong Ma, and Liangpei Zhang.
\newblock Building damage assessment for rapid disaster response with a deep object-based semantic change detection framework: From natural disasters to man-made disasters.
\newblock {\em Remote Sensing of Environment}, 265:112636, 2021.

\bibitem{guo2021deep}
Haonan Guo, Qian Shi, Andrea Marinoni, Bo~Du, and Liangpei Zhang.
\newblock Deep building footprint update network: A semi-supervised method for updating existing building footprint from bi-temporal remote sensing images.
\newblock {\em Remote Sensing of Environment}, 264:112589, 2021.

\bibitem{lou2024integrating}
Kangkai Lou, Mengmeng Li, Fashuai Li, and Xiangtao Zheng.
\newblock Integrating local--global structural interaction using siamese graph neural network for urban land use change detection from vhr satellite images.
\newblock {\em IEEE Transactions on Geoscience and Remote Sensing}, 2024.

\bibitem{huang2018agricultural}
Yanbo Huang, Zhong-xin Chen, YU~Tao, Xiang-zhi Huang, and Xing-fa Gu.
\newblock Agricultural remote sensing big data: Management and applications.
\newblock {\em Journal of Integrative Agriculture}, 17(9):1915--1931, 2018.

\bibitem{li2020review}
Jun Li, Yanqiu Pei, Shaohua Zhao, Rulin Xiao, Xiao Sang, and Chengye Zhang.
\newblock A review of remote sensing for environmental monitoring in china.
\newblock {\em Remote Sensing}, 12(7):1130, 2020.

\bibitem{gao2024relating}
Yuhao Gao, Gensheng Pei, Mengmeng Sheng, Zeren Sun, Tao Chen, and Yazhou Yao.
\newblock Relating cnn-transformer fusion network for change detection.
\newblock {\em arXiv preprint arXiv:2407.03178}, 2024.

\bibitem{chen2024changemamba}
Hongruixuan Chen, Jian Song, Chengxi Han, Junshi Xia, and Naoto Yokoya.
\newblock Changemamba: Remote sensing change detection with spatio-temporal state space model.
\newblock {\em IEEE Transactions on Geoscience and Remote Sensing}, 2024.

\bibitem{lei2023ultralightweight}
Tao Lei, Xinzhe Geng, Hailong Ning, Zhiyong Lv, Maoguo Gong, Yaochu Jin, and Asoke~K Nandi.
\newblock Ultralightweight spatial--spectral feature cooperation network for change detection in remote sensing images.
\newblock {\em IEEE Transactions on Geoscience and Remote Sensing}, 61:1--14, 2023.

\bibitem{codegoni2023tinycd}
Andrea Codegoni, Gabriele Lombardi, and Alessandro Ferrari.
\newblock Tinycd: A (not so) deep learning model for change detection.
\newblock {\em Neural Computing and Applications}, 35(11):8471--8486, 2023.

\bibitem{chen2021remote}
Hao Chen, Zipeng Qi, and Zhenwei Shi.
\newblock Remote sensing image change detection with transformers.
\newblock {\em IEEE Transactions on Geoscience and Remote Sensing}, 60:1--14, 2021.

\bibitem{bandara2022transformer}
Wele Gedara~Chaminda Bandara and Vishal~M Patel.
\newblock A transformer-based siamese network for change detection.
\newblock In {\em IGARSS 2022-2022 IEEE International Geoscience and Remote Sensing Symposium}, pages 207--210. IEEE, 2022.

\bibitem{chen2024multi}
Huan Chen, Tingfa Xu, Zhenxiang Chen, Peifu Liu, Huiyan Bai, and Jianan Li.
\newblock Multi-scale change-aware transformer for remote sensing image change detection.
\newblock In {\em Proceedings of the 32nd ACM International Conference on Multimedia}, pages 2992--3000, 2024.

\bibitem{wang2024repvit}
Ao~Wang, Hui Chen, Zijia Lin, Jungong Han, and Guiguang Ding.
\newblock Repvit: Revisiting mobile cnn from vit perspective.
\newblock In {\em Proceedings of the IEEE/CVF Conference on Computer Vision and Pattern Recognition}, pages 15909--15920, 2024.

\bibitem{daudt2018fully}
Rodrigo~Caye Daudt, Bertr Le~Saux, and Alexandre Boulch.
\newblock Fully convolutional siamese networks for change detection.
\newblock In {\em 2018 25th IEEE international conference on image processing (ICIP)}, pages 4063--4067. IEEE, 2018.

\bibitem{fang2021snunet}
Sheng Fang, Kaiyu Li, Jinyuan Shao, and Zhe Li.
\newblock Snunet-cd: A densely connected siamese network for change detection of vhr images.
\newblock {\em IEEE Geoscience and Remote Sensing Letters}, 19:1--5, 2021.

\bibitem{zhang2020deeply}
Chenxiao Zhang, Peng Yue, Deodato Tapete, Liangcun Jiang, Boyi Shangguan, Li~Huang, and Guangchao Liu.
\newblock A deeply supervised image fusion network for change detection in high resolution bi-temporal remote sensing images.
\newblock {\em ISPRS Journal of Photogrammetry and Remote Sensing}, 166:183--200, 2020.

\bibitem{peng2020optical}
Xueli Peng, Ruofei Zhong, Zhen Li, and Qingyang Li.
\newblock Optical remote sensing image change detection based on attention mechanism and image difference.
\newblock {\em IEEE Transactions on Geoscience and Remote Sensing}, 59(9):7296--7307, 2020.

\bibitem{chen2023land}
Hongruixuan Chen, Cuiling Lan, Jian Song, Clifford Broni-Bediako, Junshi Xia, and Naoto Yokoya.
\newblock Land-cover change detection using paired openstreetmap data and optical high-resolution imagery via object-guided transformer.
\newblock {\em arXiv preprint arXiv:2310.02674}, 2023.

\bibitem{howard2017mobilenets}
Andrew~G Howard.
\newblock Mobilenets: Efficient convolutional neural networks for mobile vision applications.
\newblock {\em arXiv preprint arXiv:1704.04861}, 2017.

\bibitem{zhang2018shufflenet}
Xiangyu Zhang, Xinyu Zhou, Mengxiao Lin, and Jian Sun.
\newblock Shufflenet: An extremely efficient convolutional neural network for mobile devices.
\newblock In {\em Proceedings of the IEEE conference on computer vision and pattern recognition}, pages 6848--6856, 2018.

\bibitem{chen2023run}
Jierun Chen, Shiu-hong Kao, Hao He, Weipeng Zhuo, Song Wen, Chul-Ho Lee, and S-H~Gary Chan.
\newblock Run, don't walk: chasing higher flops for faster neural networks.
\newblock In {\em Proceedings of the IEEE/CVF conference on computer vision and pattern recognition}, pages 12021--12031, 2023.

\bibitem{liu2021swin}
Ze~Liu, Yutong Lin, Yue Cao, Han Hu, Yixuan Wei, Zheng Zhang, Stephen Lin, and Baining Guo.
\newblock Swin transformer: Hierarchical vision transformer using shifted windows.
\newblock In {\em Proceedings of the IEEE/CVF international conference on computer vision}, pages 10012--10022, 2021.

\bibitem{wang2020axial}
Huiyu Wang, Yukun Zhu, Bradley Green, Hartwig Adam, Alan Yuille, and Liang-Chieh Chen.
\newblock Axial-deeplab: Stand-alone axial-attention for panoptic segmentation.
\newblock In {\em European conference on computer vision}, pages 108--126. Springer, 2020.

\bibitem{mehta2021mobilevit}
Sachin Mehta and Mohammad Rastegari.
\newblock Mobilevit: light-weight, general-purpose, and mobile-friendly vision transformer.
\newblock {\em arXiv preprint arXiv:2110.02178}, 2021.

\bibitem{tan2019efficientnet}
Mingxing Tan and Quoc Le.
\newblock Efficientnet: Rethinking model scaling for convolutional neural networks.
\newblock In {\em International conference on machine learning}, pages 6105--6114. PMLR, 2019.

\bibitem{shi2021deeply}
Qian Shi, Mengxi Liu, Shengchen Li, Xiaoping Liu, Fei Wang, and Liangpei Zhang.
\newblock A deeply supervised attention metric-based network and an open aerial image dataset for remote sensing change detection.
\newblock {\em IEEE transactions on geoscience and remote sensing}, 60:1--16, 2021.

\bibitem{ji2018fully}
Shunping Ji, Shiqing Wei, and Meng Lu.
\newblock Fully convolutional networks for multisource building extraction from an open aerial and satellite imagery data set.
\newblock {\em IEEE Transactions on geoscience and remote sensing}, 57(1):574--586, 2018.

\bibitem{chen2020spatial}
Hao Chen and Zhenwei Shi.
\newblock A spatial-temporal attention-based method and a new dataset for remote sensing image change detection.
\newblock {\em Remote sensing}, 12(10):1662, 2020.

\bibitem{lebedev2018change}
MA~Lebedev, Yu~V Vizilter, OV~Vygolov, Vladimir~A Knyaz, and A~Yu Rubis.
\newblock Change detection in remote sensing images using conditional adversarial networks.
\newblock {\em The International Archives of the Photogrammetry, Remote Sensing and Spatial Information Sciences}, 42:565--571, 2018.

\bibitem{zhang2024bifa}
Haotian Zhang, Hao Chen, Chenyao Zhou, Keyan Chen, Chenyang Liu, Zhengxia Zou, and Zhenwei Shi.
\newblock Bifa: Remote sensing image change detection with bitemporal feature alignment.
\newblock {\em IEEE Transactions on Geoscience and Remote Sensing}, 2024.

\end{thebibliography}

\end{document}